\documentclass{bmvc2k}

\usepackage{amsmath,amssymb,amsfonts}
\usepackage{algorithmic}
\usepackage{graphicx}
\usepackage{textcomp}
\usepackage{color}
\usepackage{wrapfig}

\title{Searching for Ambiguous Objects in Videos using Relational Referring Expressions}

\addauthor{Hazan Anayurt*}{hazan.anayurt@metu.edu.tr}{1}
\addauthor{Sezai Artun Ozyegin*}{artun.ozyegin@metu.edu.tr}{1}
\addauthor{Ulfet Cetin}{ulfet.rwth@gmail.com}{1}
\addauthor{Utku Aktas}{utkuaktascs@gmail.com}{1}
\addauthor{Sinan Kalkan}{skalkan@metu.edu.tr}{1}

\addinstitution{
 Department of Computer Engineering, \\
 Middle East Technical University,\\
 Ankara, Turkey\\
 \vspace{3cm}
 * \footnotesize{Equal contribution}
}

\runninghead{Anayurt et al.}{Searching for Ambiguous Objects in Videos}

\def\etal{\emph{et al}\bmvaOneDot}

\DeclareMathOperator*{\argmax}{arg\,max}

\def\BibTeX{{\rm B\kern-.05em{\sc i\kern-.025em b}\kern-.08em
    T\kern-.1667em\lower.7ex\hbox{E}\kern-.125emX}}

\begin{document}

\maketitle

\begin{abstract}
Humans frequently use referring (identifying) expressions to refer to objects. Especially in ambiguous settings, humans prefer expressions (called relational referring expressions) that describe an object with respect to a distinguishing, unique object. Unlike studies on video object search using referring expressions, in this paper, our focus is on (i) relational referring expressions in highly ambiguous settings, and (ii) methods that can both generate and comprehend a referring expression. For this goal, we first introduce a new dataset for video object search with referring expressions that includes numerous copies of the objects, making it difficult to use non-relational expressions. Moreover, we train two baseline deep networks on this dataset, which show promising results. Finally, we propose a deep attention network that significantly outperforms the baselines on our dataset. The dataset and the codes are available at \url{https://github.com/hazananayurt/viref}.
\end{abstract}

\section{Introduction}
\begin{figure}[t]
\centerline{\includegraphics[width=0.8\textwidth]{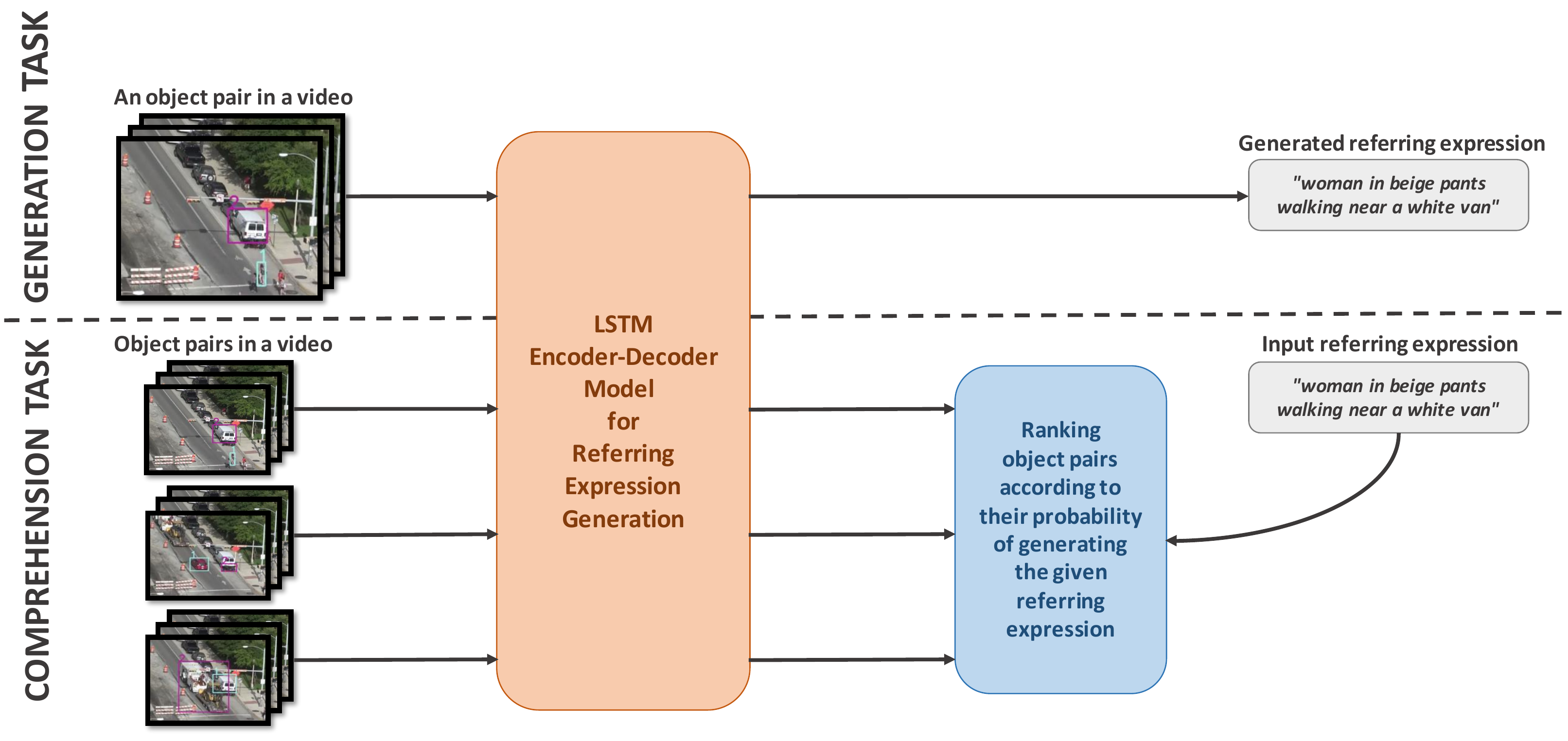}}
\caption{An overview of the generation and comprehension tasks performed by our model.}
\label{fig_inp_out}
\end{figure}

A referring expression (RE) is a description that identifies an object among many candidates. Humans are likely to use (non-relational) RE that include absolute attributes such as color or relative attributes such as size especially when there are no objects with the same attributes \cite{dale1989cooking}. However, when there are ambiguities e.g. owing to having multiple instances of the same object categories or multiple objects having the same attributes, humans prefer relational RE \cite{dale1989cooking,viethen2008use}, i.e., descriptions that identify an object with respect to other objects.

In the past few years, there have been many successful studies on REs for objects in images \cite{kazemzadeh2014referitgame,mao2016generation,nagaraja2016modeling,yu2018mattnet}. These studies generally focus on two main tasks: generation and comprehension. In a generation task, the aim is to generate an RE given an image and an object whereas, in comprehension task, the aim is to find the object given an image and an RE. 

Recently, a few studies have addressed RE comprehension for object search in videos \cite{balajee2018object,khoreva2018video,wiriyathammabhum2019referring}. However, these studies (i) only focused on the comprehension of REs, and (ii) used datasets and REs that did not necessarily require relational attributes since the level of ambiguities was low. 

In this work, we collected a dataset for REs describing an object using another (context) object. This means that every referring expression includes two objects: the main object and the context object. We collected the REs on the objects from the VIRAT (Video Image Retrieval and Analysis Tool) dataset \cite{oh2011large}, which is a surveillance video dataset mainly containing people and vehicles, and the subset of the ILSVRC2015 dataset \cite{ILSVRC15} which contains vehicles.

Using the dataset we collected, we were able to train a model that recognizes the relationship between two objects including the classes and physical appearances of the objects and the actions they are performing. To achieve this, we used a model similar to the one used in \cite{mao2016generation} for RE comprehension and generation on images. The idea behind our model is first training an LSTM generator that, given two objects in a video, generates an RE describing the first object using the second object. Then, we can use this LSTM generator in the comprehension task. See Fig. \ref{fig_inp_out} for an illustration.

{\noindent}\textbf{Related Work on Referring to Objects in Images:} With advances in deep learning, it has been possible to link language and vision with superior performance than before. One of the first to address such a link for referring to objects in images with textual descriptions is Kazemzadeh \etal \cite{kazemzadeh2014referitgame}, who collected a large-scale dataset using a game-like activity. Many following studies extended and proposed better ways to use referring expressions. E.g., Mao \etal \cite{mao2016generation} demonstrated that RE comprehension and generation can be addressed by a single model thanks to the generative and discriminative capabilities of a Recurrent Neural Network. In their model, an RE is generated using an LSTM \cite{hochreiter1997long} network that is initialized with VGG features \cite{simonyan2014very} extracted from the input image. For comprehension, the same LSTM network is used to evaluate the compatibility with the features of the input image. 

Nagaraja \etal \cite{nagaraja2016modeling} came up with another interesting approach by formulating RE generation and comprehension as a Multiple Instance Learning problem for learning RE generation with respect to a context object even when such objects are not labeled in the dataset. They do this by creating positive and negative bags from object pairs. The positive bags contain the referred object as the main object while the negative bags do not contain the referred object at all. This way, there is no possibility that the referred pair is in the negative bag whereas it could be any of the pairs in the positive bag but which object is the context object is not known. They claim that, by modelling the context between objects, they were able to achieve a better performance than when they used only the main object's properties.

Similar to other problems linking computer vision and language, it has recently been shown that an attention capability improves the performance of RE comprehension \cite{yu2018mattnet}. This is a ``module''-level attention model that modulates contributions from location, object and action/relation. 

{\noindent}\textbf{Related Work on Referring to Objects in Videos:} Inspired from similar studies on images, recently, many have focused on using REs for objects in videos \cite{balajee2018object,khoreva2018video,wiriyathammabhum2019referring}. These studies have shown that REs in videos are more challenging since they also contain temporal information, and the dimensionality of the inputs is much higher. 

Vasudevan \etal \cite{balajee2018object} have studied linking human gaze with REs. In their work, the gaze location is estimated from the user's eyes, and an object corresponding to an uttered RE is sought in a region around the gaze location. A study done by Khoreva \etal \cite{khoreva2018video} has addressed the segmentation of a referred object in a video given an RE. They have evaluated their methods on video object segmentation datasets. In a more recent study by Wiriyathammabhum \etal \cite{wiriyathammabhum2019referring}, the MattNet model \cite{yu2018mattnet} was extended for REs in videos by including motion modules and attention on motion modules.

{\noindent}\textbf{Our Contributions:} Looking at the related studies, we can summarize our contributions as follows:

{\noindent}\textbf{(1) A Dataset with Ambiguities}: We introduce a new dataset that includes REs on pairs of objects. The dataset is highly ambiguous, including numerous instances of the same objects, therefore, relational REs are required.
    
The three studies \cite{balajee2018object,khoreva2018video,wiriyathammabhum2019referring} on using REs for objects in videos have introduced their own datasets since they all addressed a different setting and application for REs. However, these datasets are not suitable for our goals since (i) they require additional modalities like gaze or segmentation, and (ii) ambiguity in these datasets is limited and therefore, they do not focus on relational REs. 
    
{\noindent}\textbf{(2) A Deep Attention Model} that can relate a relational RE with an object pair. Unlike other studies on REs on videos, our model can be used for both comprehension and generation tasks. Moreover, we provide two baseline models that are extensions of REs for objects in images (namely, \cite{mao2016generation}) to REs for objects in videos.

\section{A Dataset of Relational Referring Expressions (REs) for Objects in Videos}

For our dataset, we used videos from two video datasets: The VIRAT Ground \cite{oh2011large} and ILSVRC2015 \cite{ILSVRC15} datasets. Our dataset consists of videos and bounding boxes from these two datasets and REs we collected for object pairs in these videos. VIRAT is a video surveillance dataset that contains videos of places like streets and parking lots. Due to this, the main object classes are people and vehicles. Because we wanted to narrow our domain down to fit the majority of classes in VIRAT, we only used videos that contained vehicles from the ILSVRC dataset. Due to this, VIRAT makes up most of our videos. Other than this, we did not put a restriction on object categories or REs on them, which means that our dataset contains objects (and REs using these objects) such as garbage bins, traffic cones and objects carried by people e.g. bags or boxes.

One thing we did put a restriction on while choosing object pairs was whether the pair actually had a relation or not. This was because an RE defining one object using the other could not be written if the objects did not have any meaningful ``overlap'' in the video. We saw that the number of pairs we obtained were still too high even after eliminating pairs that were too far apart spatially or temporally; therefore, we visually observed each pair to decide manually whether or not a meaningful RE could be written for them. The REs we have collected for this dataset are only for the pairs that remained after both eliminations.

To collect the REs for our dataset, we used the Microworkers crowdsourcing platform. The video sequences we showed the workers were only the parts of a video starting when the first object of the pair to enter the video enters and ending when the last one of the pair to exit the video exits. Each video sequence contained an object pair with one object labeled as the first object and the other object labelled as the second and we asked for the workers to provide an RE for the first object using the second object as the context object after watching the video. Therefore, the order of the labels was important. We also reversed the order of the pair, and collected REs for the reverse order as well.

We collected six REs each object pair (three for the straight and three for the reverse order). We asked the workers to give as much detail as possible and we did not have any vocabulary or grammar restrictions apart from asking the answers to be noun phrases defining the main object w.r.t. the context object without using full sentences. Differently from the previous datasets, we did not require for the REs to be unambiguous; hence, one RE could be correctly identifying more than one pair.

\begin{figure*}[]
\centerline{\includegraphics[width=1\textwidth]{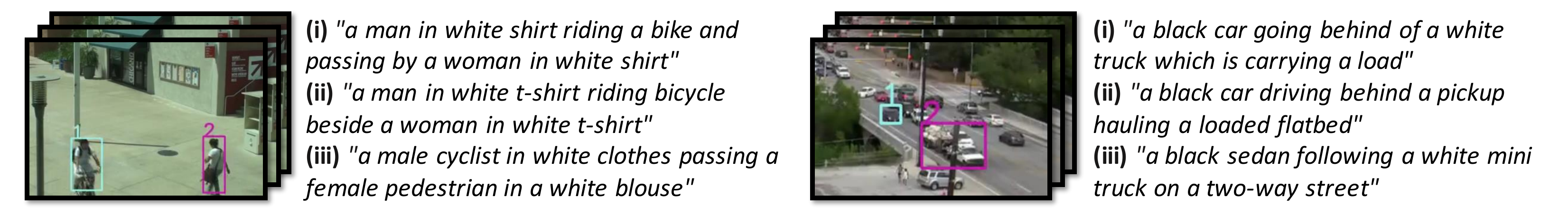}}
\caption{Two object pairs from our dataset and the three REs collected for them.}
\label{fig:dataset_examples}
\end{figure*}

Two example object pairs from our dataset along with the collected REs can be seen in Fig. \ref{fig:dataset_examples}. In Tables \ref{tab:video_statistics} and \ref{tab:refexp_statistics} we provide some descriptive statistics about our dataset. REs with length greater than 25 are not used during training as explained in Section \ref{sect:train_and_impl}. Moreover, words that occur only once are removed from our vocabulary.

\begin{table}
\small
\parbox[t]{.45\linewidth}{
\centering
\caption{Information about the videos in our dataset. \label{tab:video_statistics}}
\footnotesize
    \begin{tabular}{| l | c | c | c |}\hline
        & {\footnotesize \textbf{Avg.}} & {\footnotesize \textbf{Min.}} & {\footnotesize \textbf{Max.}}  \\  \hline \hline
        Sequence length (sec) & 19.00 & 1 & 42 \\  \hline
        $\#$ objects per video & 9.38 & 2 & 46 \\  \hline
        $\#$ pairs per video & 19.57 & 2 & 148 \\ \hline \hline
        $\#$ videos & \multicolumn{3}{c|}{125 ({\scriptsize VIRAT}) + 37 ({\scriptsize ILSVRC})} \\ \hline
        $\#$ objects (total) & \multicolumn{3}{c|}{1520} \\ \hline
        $\#$ pairs (total) & \multicolumn{3}{c|}{3170} \\ \hline
    \end{tabular}
}
\hfill
\parbox[t]{.45\linewidth}{
\centering
\caption{Information about the REs in our dataset. $^\bullet$ stands for the number of words in an RE, and $\star$ denotes the number of occurences of words in the dataset. \label{tab:refexp_statistics}}
\footnotesize 
    \begin{tabular}{| l | c | c | c |}\hline
        & {\footnotesize \textbf{Avg.}} & {\footnotesize \textbf{Min.}} & {\footnotesize \textbf{Max.}}  \\  \hline \hline
        RE length & 11.70 & 5 & 37 \\  \hline
        $\#$ RE per object & 12.51 & 6 & 90 \\  \hline \hline
        $\#$ RE & \multicolumn{3}{c|}{9416 ($\bullet < 25$) + 94 ($\bullet \ge 25$)} \\ \hline
        $\#$ words in REs & \multicolumn{3}{c|}{708 ($\star>1$) + 370 ($\star=1$)} \\ \hline
    \end{tabular}
}
\end{table}

\section{Video Object Search using Relational Referring Expressions (VIREF)}

For our model, similar to \cite{balajee2018object,khoreva2018video,wiriyathammabhum2019referring}, we assume that objects have been already detected or annotated. Then, given two objects in a video, our goal is to generate the most probable referring expression defining the first object using the second object, which is called the generation task. We also use the model trained on the generation task to find the most suitable object pair among labeled objects in a video given an RE as input, which is the comprehension task.

We use a method similar to \cite{mao2016generation} and approach the generation and comprehension tasks together. We first encode the objects in videos using an LSTM, then we feed the encoded object pair to a decoder LSTM. We use the hidden states of the decoder to calculate the attention weights of the objects' features, enabling our model to attend to different combinations of features according to the word it wants to predict. After the attention, we combine the encoded features that the model has attended to and the hidden state of the decoder to calculate the probability of each word in our vocabulary to try to obtain the most probable RE given an object pair in a video. We also use these probabilities while searching for objects in a video (comprehension task). An overview of our model is provided in Fig. \ref{fig:viref_model}.

\subsection{Generation Task}
\label{sect:generation_task}

The generation task can be formally defined as finding the most likely sequence of words, $\mathbf{r}=:<w_1, ..., w_n>$, (i.e., the referring expression) from the vocabulary given a video sequence $\mathbf{v}=:<I_1, ..., I_m>$ ($I_i$ is the $i^{th}$ frame) with annotated boxes for the main object, $\mathbf{o}^t=:<B^t_1, ..., B^t_m>$, and the context object, $\mathbf{o}^c=:<B^c_1, ..., B^c_m>$. $B$ is simply a vector with the coordinates of the top-left and the bottom-right corners of the bounding box.

For the generation task, we use an LSTM decoder architecture similar to \cite{mao2016generation} with two main differences, the first of which is using an encoder LSTM on the object features. This helps in adapting the model to the video domain by allowing it to capture the temporal relation between the objects more accurately. The second one is the attention module attached on top of the outputs of the decoder.

The input ($\mathbf{x}_i$) of the encoder LSTM at step $i$ are the features extracted from the $i^{th}$ frame of the input video (sampled each second). Denoting the VGG16 (fc1) \cite{simonyan2014very} features by $\phi()$, $\mathbf{x}_i$ is a concatenation of the following:
\begin{equation}
    \mathbf{x}_i = <\phi(I_i(B^t_i)),\ \phi(I_i(B^c_i)),\  \phi(I_i),\ \phi(M(B^t_i)),\ \phi(M(B^c_i))>,
\end{equation}
\noindent where $I_i(B)$ denotes an image with the pixels in bounding box $B$; and $M(B)$ is a binary image (with the same size as $I_i$) where the pixels are white inside $B$ and black elsewhere.

The hidden state of the encoder LSTM at step (frame) $i$ can be denoted as follows:
\begin{equation}
    \mathbf{h}_i^e = LSTM^e (\mathbf{x}_i^{a_0},\ h_{i-1}^e),
\end{equation}
\noindent where $LSTM^e$ is the encoder LSTM, $\mathbf{x}_{i}^{a_0}$ is the $i^{th}$ input scaled with the initial attention weights $a_0=<a_0^1,\ a_0^2,\ a_0^3,\ a_0^4,\ a_0^5> \in \mathbb{R}^5$ and $\mathbf{h}_{i}^e$ is the encoder hidden state at step $i$. $\mathbf{h}_{0}^{e}$ is a trainable parameter and it is initialized as 0. The attention weights are integrated as follows:
\begin{equation}\label{eq:scale}
    \mathbf{x}_i^{a_0} = <a_0^1\times \phi(I_i(B^t_i)),\ a_0^2\times \phi(I_i(B^c_i)),\ a_0^3\times \phi(I_i),\  a_0^4\times \phi(M(B^t_i)),\ a_0^5\times \phi(M(B^c_i))>.
\end{equation}

The rest of our model resembles a regular image captioning architecture with the addition of a feature attention mechanism. As input to the decoder, we used GloVe \cite{pennington2014glove} 50-dimensional vector representations of the words. The hidden state of the decoder at step $j$ can be denoted as follows:
\begin{equation}
    \mathbf{h}_{j}^{d} = LSTM^{d} (\psi(w_{j}),\  h_{j-1}^{d}),
\end{equation}
\noindent where $w_j$ is the word at step $j$ and $\psi(\cdot)$ is the function that takes a word to its embedding. The last hidden state of the encoder is fed into the decoder as its initial hidden state, i.e., $\mathbf{h}_1^{d} = LSTM^{d} (\psi(w_{start}), h_{m}^{e})$, where $\psi(w_{start})$ is the embedding vector of a pre-defined start token.

\begin{figure}
        \centering
    \begin{minipage}{0.53\textwidth}
        \centering
        \includegraphics[width=0.78\textwidth]{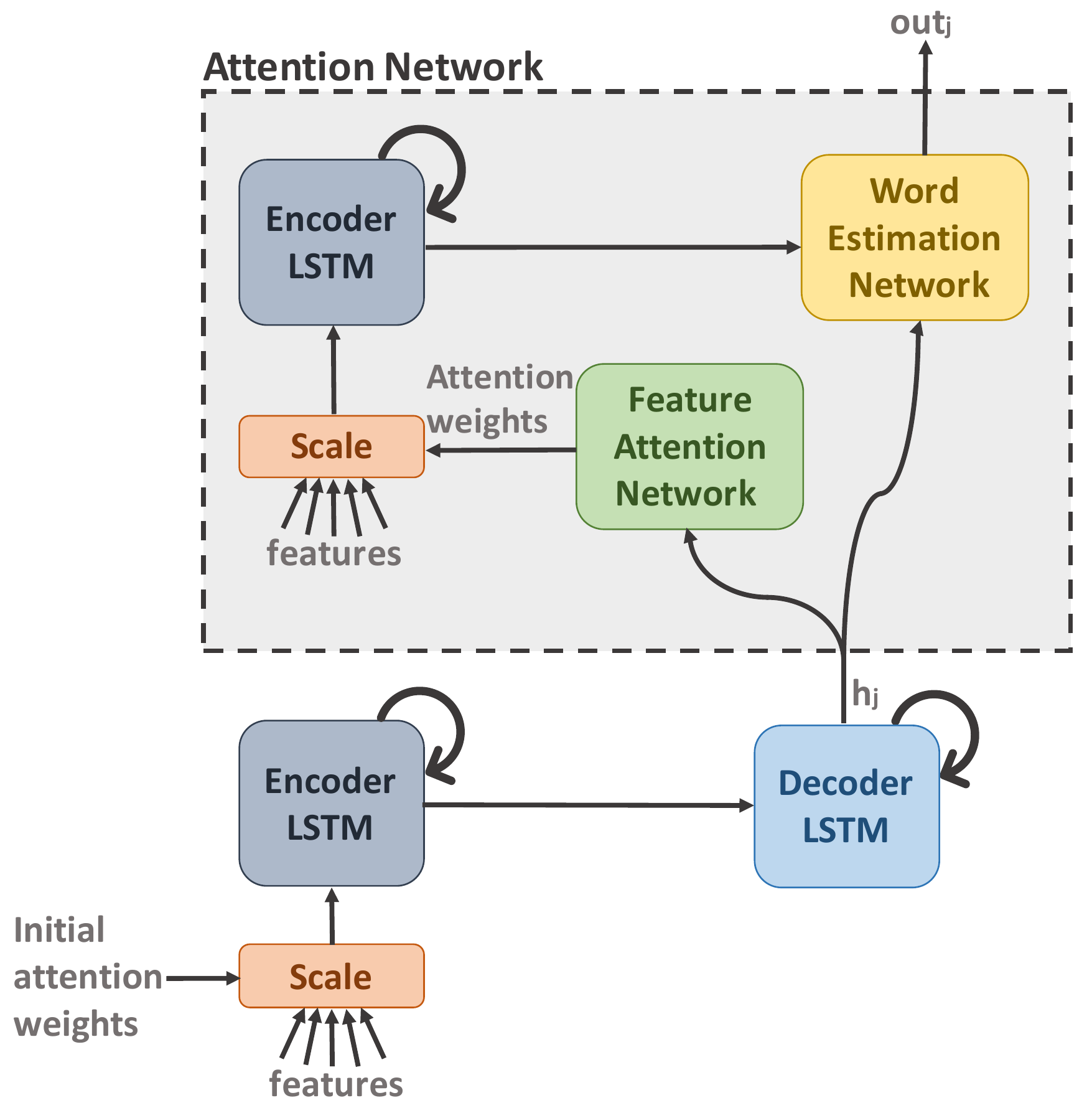}
        \caption{The VIREF model. \label{fig:viref_model}}
    \end{minipage}\hfill
    \begin{minipage}{0.47\textwidth}
        \centering
         \includegraphics[width=0.44\textwidth]{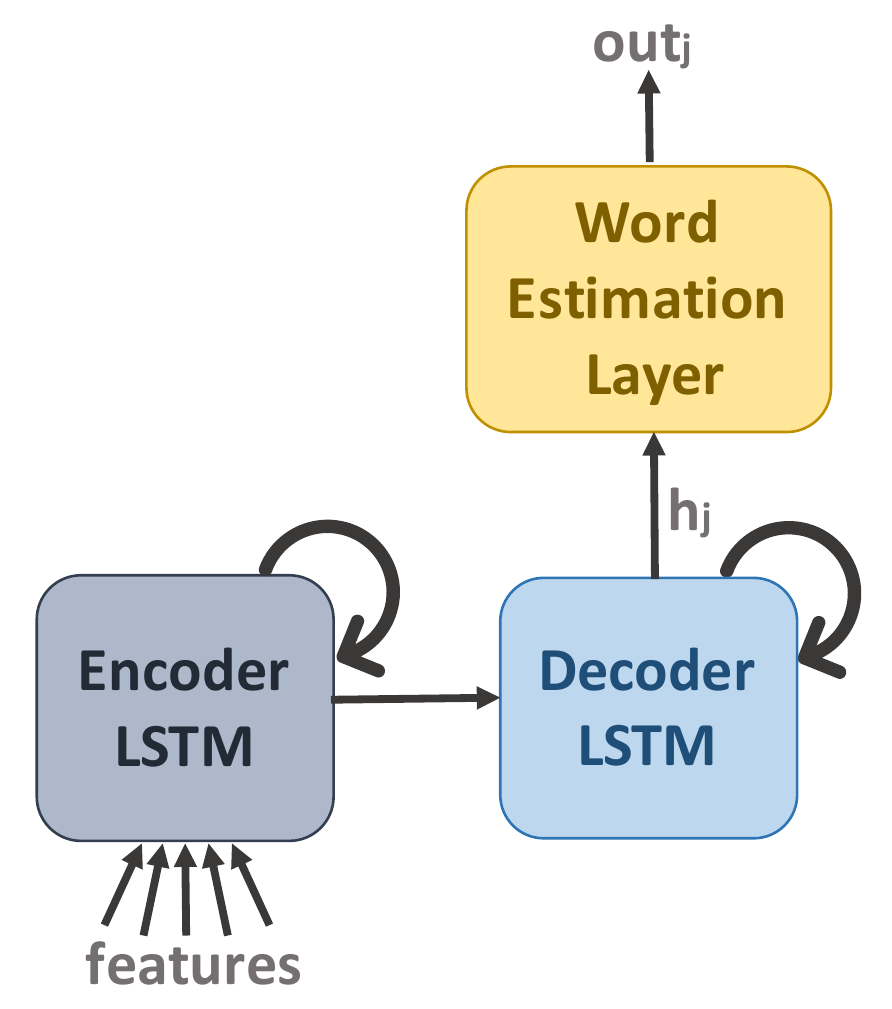}
        \caption{The VIREF-a model (VIREF without attention). \label{fig:viref_a_model}}
        \includegraphics[width=0.53\textwidth]{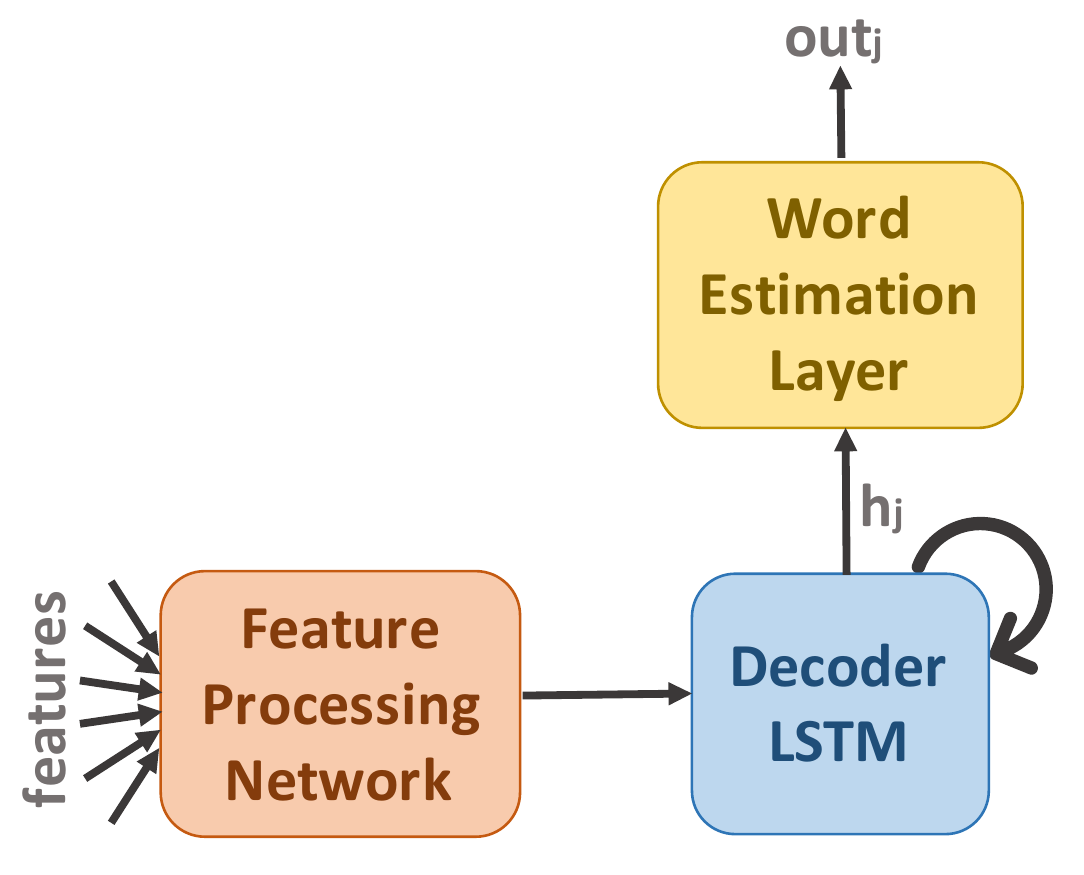}
        \caption{The VIREF-e model (VIREF without LSTM encoder). \label{fig:viref_e_model}}
    \end{minipage}
\end{figure}

At each time step $j$ of the decoder, the attention weight vector $a_j$ is calculated using a network named Feature Attention Network ($FAN$) as $a_j = FAN(h_j^d)$. $a_j$ is used as the attention weights to the features of the encoder (Fig. \ref{fig:viref_model}). The attention weights $a_j$ are integrated into the inputs in the same way as shown for $a_0$ in Eq. \ref{eq:scale}. Then, with those weights, the encoder is run again. Combining the last hidden state of the newly run encoder and the $\mathbf{h}_j^d$, the output is calculated using the Word Estimation Network ($WEN$) as follows:
\begin{equation}\label{eq:output}
    \textbf{out}_j = WEN(h_j^d,\ LSTM^e(x_m^{a_j},\  LSTM^e(x_{m-1}^{a_j},\ \hdots))).
\end{equation}
The output of $WEN$ is passed through Softmax to obtain a probability distribution over the words in our vocabulary. With this, we obtain the probability of an RE, $p(\mathbf{r}\ |\ \mathbf{v}, \mathbf{o}^t, \mathbf{o}^c)$, by multiplying the word probabilities. In test time, we sample the most probable RE (i.e., $\argmax_{\mathbf{r}} p(\mathbf{r}\ |\ \mathbf{v}, \mathbf{o}^t, \mathbf{o}^c)$) using beam search with beam size 3, similar to \cite{mao2016generation}. The illustration of the VIREF generation model can be seen in Fig. \ref{fig:viref_model}.

The attention mechanism we employ is similar to the modular attention mechanism employed in \cite{wiriyathammabhum2019referring, yu2018mattnet}. The main differences are that attention in our model is performed over the extracted features at each time step, not over the modules that summarize information over time separately, and it is adapted to video domain by using an encoder LSTM. 

As can be seen in Eq. \ref{eq:output}, the encoder is run again for each time step of the decoder with its corresponding feature attention weights. It may be argued that this is too much computational overhead; however, in this way, the decoder can decide which input features it would attend to according to the word it wants to predict and this allows the features to interact in the encoding step. If we wanted to have feature-level attention but only ran the encoder once and attended to the encoded features, we would have to encode each feature separately. This would prevent the encoder from capturing the relation between the features such as an object and its location, or the motion of the objects with respect to each other.

\subsection{Comprehension Task}
\label{sect:comprehension_task}

For the comprehension task, we do not train a separate model but use the model we trained for the generation task. This becomes possible if we assume uniform prior for the probability of the video and the places of the objects in it (meaning they can occur anywhere in any video). Under this assumption, using the Bayes' Rule, the probability of a video and an object pair given a referring expression becomes directly proportional to the probability of a referring expression given an object pair since the evidences are the same:
\begin{equation}
\argmax_{\mathbf{v}, \mathbf{o^t}, \mathbf{o^c}} p(\mathbf{v}, \mathbf{o^t}, \mathbf{o^c}| \mathbf{r}) = \argmax_{\mathbf{v}, \mathbf{o^t}, \mathbf{o^c}} p(\mathbf{r} | \mathbf{v}, \mathbf{o^t}, \mathbf{o^c}).
\end{equation}
Therefore, when an RE is taken as input, we calculate the probability of that RE for each object pair in each video, $p(\mathbf{r} | \mathbf{v}, \mathbf{o^t}, \mathbf{o^c})$,  using the generator model as explained in Section \ref{sect:generation_task}. Then, we rank the object pairs according to the probability of them giving the input RE. Mao \etal \cite{mao2016generation} have used a similar approach.

\subsection{Architecture and Loss}
\label{sect:arch_and_loss}

For both encoder and decoder, we use a six-layer LSTM, whereas Feature Attention Network and Word Estimation Network both consist of 3 fully-connected layers. Both the LSTMs and the fully-connected networks in the attention module use Dropout with dropping probability of $0.2$. The initial attention weights are trainable parameters and all attention weights are used after being passed through the Softmax function. The input features are $4096$-dimensional each and add up to $20480$ when concatenated after scaling(see Eq. \ref{eq:scale}). The output of the Word Estimation Network is of size $1024$, which is the size of our vocabulary, and is also passed through Softmax. For training the networks, we use cross-entropy loss.

\section{Baseline Models}

To compare our method to, we provide two baseline models. The architectures are similar to VIREF in certain parts with the main differences being in the encoder module. Essentially, these models are simplified versions of the VIREF model to show the performance improvements provided in both the generation and comprehension tasks by the LSTM encoder and attention modules. 

{\noindent}\textbf{VIREF without Attention (VIREF-a): } VIREF-a is the same model as VIREF without the extra attention module, as shown in Fig. \ref{fig:viref_a_model}. This model can be described as a straightforward extension of the model used in \cite{mao2016generation} to the video domain, without the usage of video-related feature extractors. The feature extractors in image domain are still more powerful than the ones in the video domain, therefore, to efficiently use the information in the time dimension, an LSTM encoder is utilized.

{\noindent}\textbf{VIREF without LSTM encoder (VIREF-e)} Another straightforward extension of \cite{mao2016generation} is directly using video feature extractors and feeding them into decoder LSTM (as illustrated in Fig. \ref{fig:viref_e_model}). Using this kind of approach would remove the necessity of using a recurrent network to capture the temporal information.

We used six features that could replace the LSTM-encoded VGG16 features. The first two are the averages of VGG16 (fc1) \cite{simonyan2014very}  features for the main and the context objects over time. The other four are the C3D (fc6) \cite{tran2015learning} features for the main object, the context object, and the main \& context objects together (formed by blackening regions outside of their bounding boxes), and the whole scene. We believe that VGG16 features of the objects could capture the fine details of the objects and C3D features could contain motion related information.

The model takes these six features, processes them using fully-connected layers and uses the same decoder as VIREF-a. The only difference is that encoder part uses different kind of feature extraction methods instead of the LSTM encoder used in the other two models. Each of the features is of the size $4096$ and the features add up to $24576$ when concatenated. After the concatenation, features are passed through Feature Processing Network which consists of 3 fully-connected layers.

\section{Experiments}
In this section, we evaluate our method and compare it against the baselines on generation and comprehension tasks.

\subsection{Training and Implementation Details}
\label{sect:train_and_impl}

We split our dataset into training, validation and test sets respectively with $60\%$, $10\%$, $30\%$ percentages. For training the networks, we used Adam optimizer with a learning rate of $10^{-4}$ and a batch size of 10. We used the validation set to perform early-stopping and observed this to be a sufficient measure against overfitting in our experiments.

The decoder LSTMs output a probability distribution over the words in our vocabulary. The vocabulary includes (i) the words that appear at least twice in our training set (634 words), and (ii) the words most frequently used in the Google RefExp dataset \cite{mao2016generation} ($386$ words). In total, with <\textit{start}>, <\textit{end}>, <\textit{unk}> and <\textit{nil}> tokens, our vocabulary contains $1024$ words.

During training, we use padding (the <\textit{nil}> token) so that the referring expressions will have the same length. However, the number of very long REs only make up a very small percentage of the dataset and to be able to use less padding for evening out the length of all REs, we exclude every RE that has a length of 25 or greater in training. Out of the $9,510$ REs we collected in total, only 94 exceed this length constraint and we are able to avoid higher training times at the cost of only $1\%$ of our data.

\subsection{Generation Results}

\begin{figure}[]
\centerline{\includegraphics[width=1\textwidth]{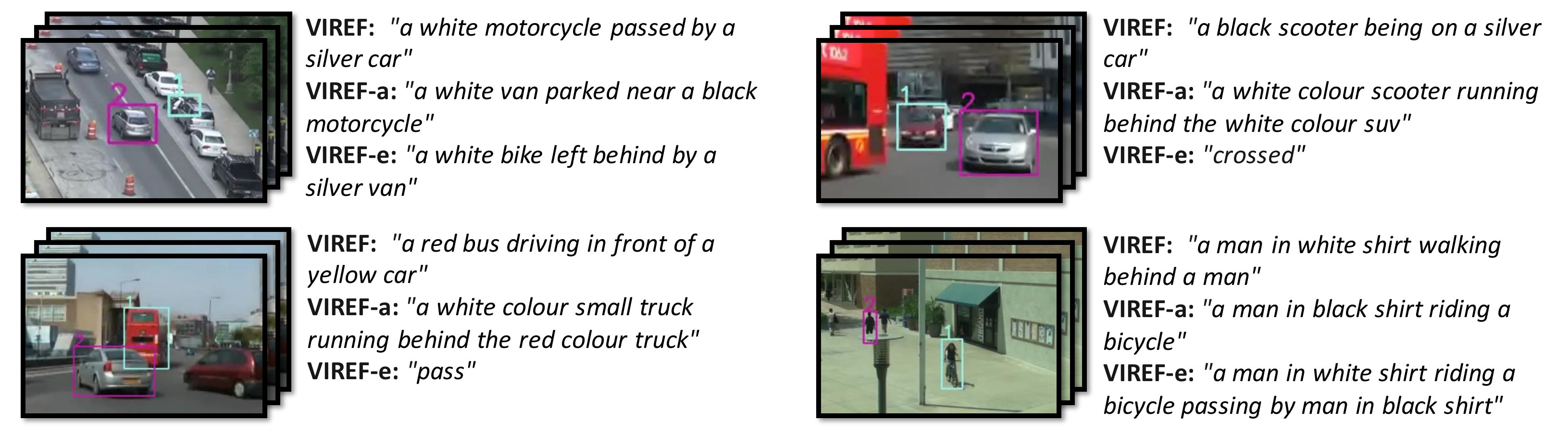}}
\caption{Sample generated REs. Object with label 1 is the main object, and the one with label 2 is the context object.} 
\label{fig:generation_results}
\end{figure}

To evaluate the results of the generator, we calculated average scores using the BLEU \cite{papineni2002bleu} and METEOR \cite{lavie2007meteor} metrics. The averaged BLEU-4 and METEOR scores of the models on the test set can be seen in Table \ref{tab:generation_scores}. We observe that VIREF outperforms both baseline models in terms of the BLEU and METEOR metrics. However, even though VIREF is the best model in terms of BLEU and METEOR scores, it can be seen that it generates REs from a more narrow set of words compared to VIREF-e. According to our observations, VIREF-e's vast usage of different words is sometimes accompanied by the generation of REs which do not meaningfully describe an object. We believe that this has affected its average BLEU and METEOR scores. Some of the generation results of the models can be observed in Fig. \ref{fig:generation_results}.

\begin{table}[]
    \centering
    \caption{RE generation performance of the models. Higher is better. \label{tab:generation_scores}}
    \footnotesize
    \begin{tabular}{|c|c|c|c|}\hline
        \textbf{Method} & \textbf{Average BLEU-4 Score} & \textbf{Average METEOR Score} & \textbf{\# of words used in the output} \\ \hline\hline
        VIREF-e & 0.1197 & 0.4054 & \textbf{233} \\ \hline
        VIREF-a & 0.1498 & 0.4580 & 75 \\ \hline
        VIREF & \textbf{0.2365} & \textbf{0.5492} & 91 \\ \hline
    \end{tabular}
\end{table}

\subsection{Comprehension Results}

The comprehension part of the model, as mentioned in Section \ref{sect:comprehension_task}, uses the generator model. To evaluate the comprehension model, for each video, we selected an RE from the test set and calculated a matching score with every pair in that video. We assumed that only the object pair it was written for is the correct answer. This may have, however, led to a score that is less than the actual score, since our dataset consists of ambiguous examples and an RE may actually represent more than one object pair.

We used two different retrieval task evaluation measures: mean average precision (mAP) and rank-$k$ accuracy. AP is defined simply as $1/k$ if the correct pair is retrieved at rank $k$ (since we have only a single ``relevant document'', average precision is directly equal to precision). On the other hand, rank-$k$ accuracy is the percentage of results in which the correct pair is retrieved at top $k$ object pairs.

\begin{figure}[]
\centerline{\includegraphics[width=1\textwidth]{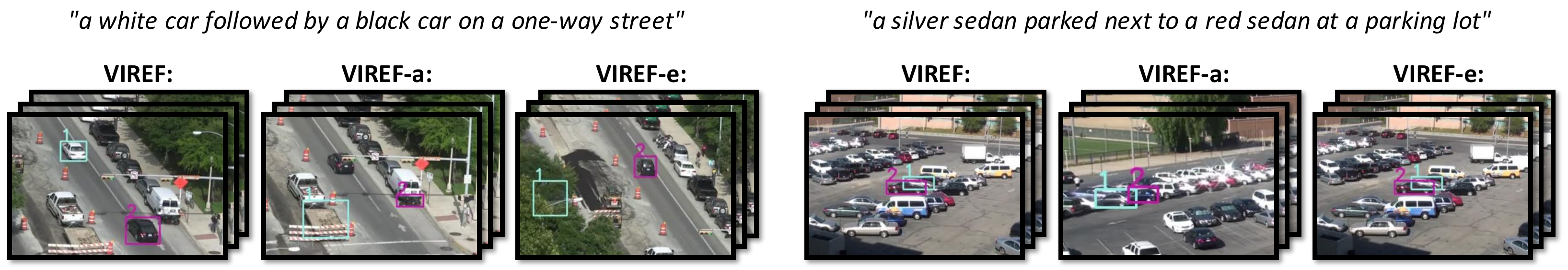}}
\caption{Sample comprehension results. Object with label 1 is the main object, and the one with label 2 is the context object.} 
\label{fig:comprehension_results}
\end{figure}

Table \ref{tab:comprehension_scores} lists the comprehension results of the methods in terms of mAP and rank-1, rank-2, and rank-3 accuracies. We observe that VIREF again performs better than the other models, thanks to its attention model. See also Fig. \ref{fig:comprehension_results} for some sample results.

\begin{table}[]
    \centering
    \caption{RE comprehension results. Higher is better. \label{tab:comprehension_scores}}
    \footnotesize
    \begin{tabular}{|c|c|c|c|c|}\hline
        \textbf{Method} & \textbf{mAP} & \textbf{rank-1 accuracy} & \textbf{rank-2 accuracy} & \textbf{rank-3 accuracy}\\ \hline\hline
        VIREF-e & 0.55 & 0.35 & 0.61 & 0.69 \\ \hline
        VIREF-a & 0.46 & 0.26 & 0.49 & 0.57 \\ \hline
        VIREF & \textbf{0.65} & \textbf{0.47} & \textbf{0.69} & \textbf{0.78} \\ \hline
    \end{tabular}
\end{table}

\subsection{Size and Running-time of the Models}

Table \ref{tab:runtime_analysis} lists the number of parameters and the GPU time used by the models. We can see that VIREF-e has the most number of parameters since the encoding part is a fully-connected network. However, this provides a gain in terms of running speed.

\begin{table}[h]
    \centering
    \caption{\# params. and speed (avg. of 100 samples on Nvidia GeForce GTX 1060). \label{tab:runtime_analysis}}
    \footnotesize
    \begin{tabular}{|c|c|c|c|}\hline
        \textbf{Method} & \textbf{$\#$ of parameters} & \textbf{Generation time (sec)} & \textbf{Comprehension time (sec)}\\ \hline\hline
        VIREF-e & 85.8M & 0.402 & 0.007 \\ \hline
        VIREF-a & 27.1M & 0.434 & 0.014 \\ \hline
        VIREF & 28.2M & 1.403 & 0.252 \\ \hline
    \end{tabular}
\end{table}

\section{Conclusion}

In this work, we addressed linking objects in videos with relational referring expressions (REs). For this, we first collected a dataset of REs for a subset of videos from VIRAT and ILSVRC datasets. The videos we specifically have chosen included highly ambiguous settings with numerous occurrences of objects.

Moreover, we proposed an encoder-decoder recurrent architecture with which we can both comprehend an RE (i.e. identify the matching object pair in a video) and generate an RE identifying a pair of objects in a video. At each decoding stage, our model can attend to the features in the encoding stage. Compared with the two baselines that we have developed, our model performs significantly better in both generation and comprehension tasks.

\section*{Acknowledgments}
This work was partially supported by the Scientific and Technological Research Council of Turkey (T\"UB\.ITAK) through the project titled ``Object Detection in Videos with Deep Neural Networks'' (project no 117E054). The authors would like to thank Dr. Emre Akbas for discussions and suggestions on the paper.

\bibliography{references.bib}

\end{document}